\documentclass[runningheads]{llncs}
\usepackage[T1]{fontenc}
\usepackage{graphicx}
\usepackage{booktabs}
\usepackage{multirow}
\usepackage{amsmath,amssymb,amsfonts}
\usepackage{caption}
\usepackage{xcolor}
\usepackage[misc]{ifsym} 
\usepackage[colorlinks=true,citecolor=blue,urlcolor=blue]{hyperref}
\usepackage{bbding}

\begin{document}
\title{Cross Prompting Consistency with Segment Anything Model for Semi-supervised \\Medical Image Segmentation}

\titlerunning{Cross Prompting Consistency with Segment Anything Model}
%
%\titlerunning{Abbreviated paper title}
% If the paper title is too long for the running head, you can set
% an abbreviated paper title here
%
\author{Juzheng Miao\inst{1} \and
Cheng Chen\inst{2}\textsuperscript{(\Letter)} \and
Keli Zhang\inst{3} \and Jie Chuai\inst{3} \and \\Quanzheng Li\inst{2,4} \and Pheng-Ann Heng\inst{1,5}}
% index{Miao, Juzheng}
% index{Chen, Cheng}
% index{Zhang, Keli}
% index{Chuai, Jie}
% index{Li, Quanzheng}
% index{Heng, Pheng-Ann}
%
\authorrunning{J. Miao et al.}
% First names are abbreviated in the running head.
% If there are more than two authors, 'et al.' is used.
%
\institute{Department of Computer Science and Engineering,\\
The Chinese University of Hong Kong, Hong Kong, China \and
Center for Advanced Medical Computing and Analysis, Massachusetts General\\
Hospital and Harvard Medical School, Boston, MA, USA\\
\email{cchen101@mgh.harvard.edu} \and
Huawei Noah’s Ark Lab, Shenzhen, China \and
Data Science Office, Massachusetts General Brigham, Boston, MA, USA \and
Institute of Medical Intelligence and XR,\\
The Chinese University of Hong Kong, Hong Kong, China
}
\maketitle              % typeset the header of the contribution
\begin{abstract}
Semi-supervised learning (SSL) has achieved notable progress in medical image segmentation. To achieve effective SSL, a model needs to be able to efficiently learn from limited labeled data and effectively exploit knowledge from abundant unlabeled data. 
Recent developments in visual foundation models, such as the Segment Anything Model (SAM), have demonstrated remarkable adaptability with improved sample efficiency.
To harness the power of foundation models for application in SSL, we propose a cross prompting consistency method with segment anything model (CPC-SAM) for semi-supervised medical image segmentation.
Our method employs SAM's unique prompt design and innovates a cross-prompting strategy within a dual-branch framework to automatically generate prompts and supervisions across two decoder branches, enabling effectively learning from both scarce labeled and valuable unlabeled data. 
We further design a novel prompt consistency regularization, to reduce the prompt position sensitivity and to enhance the output invariance under different prompts. 
We validate our method on two medical image segmentation tasks.
The extensive experiments with different labeled-data ratios and modalities demonstrate the superiority of our proposed method over the state-of-the-art SSL methods, with more than 9\% Dice improvement on the breast cancer segmentation task.
% Our code will be made publicly available.
Code is available at: \url{https://github.com/JuzhengMiao/CPC-SAM}.

\end{abstract}
\section{Introduction}
Segmentation is an essential step for accurate disease diagnosis and treatment planning~\cite{bernard2018deep,ma2024segment}.
Although deep learning methods have obtained impressive results in various organ or lesion segmentation tasks~\cite{isensee2021nnu}, a large scale of labeled data is required, which are extremely expensive and time-consuming to collect.
Given that unlabeled data is typically plentiful in practice, semi-supervised learning (SSL) emerges as a compelling approach by efficiently leveraging both the limited labeled data and the extensive amounts of unlabeled data~\cite{bai2017semi,fan2020inf,wu2022mutual,yang2023revisiting,yu2019uncertainty}.
% Current semi-supervised segmentation methods can be mainly divided into two categories, i.e., self-training methods by using pseudo-labels as supervision for unlabeled images~\cite{bai2017semi,fan2020inf} and consistency regularization methods by applying consistency constraints on different predictions~\cite{wu2022mutual,yang2023revisiting,yu2019uncertainty}.
% Therefore, semi-supervised learning (SSL) methods for medical image segmentation has been widely investigated and achieved promising results, using limited labels and plenty of unlabeled data on top of a self-training~\cite{bai2017semi,fan2020inf} or consistency regularization~\cite{wu2022mutual,yang2023revisiting,yu2019uncertainty} framework.

%对于SSL，成功的关键在于1. 打基础：从有限的labeled data 中efficiently学到足够general的knowledge for segmentation；2. 在前一个基础上，在unlabeled data就可以有比较好的prediction，那么就可以xxx。之前的工作主要关注到2。针对1的提升，我们就关注到了SAM，因为xxx，然后在此基础上，我们就可以针对2xxx。之前SAM相关的工作主要需要用到fully labeled xxx，有xxx的问题。very recently也有limited work开始关注sam在semi情况下xxx。比如xxx方法xxx。但是这些方法有xxx的问题。

%第三段，开始讲motivation。比如为什么从prompt着手 xxx

We consider the keys to the success of SSL methods are two folds. 
First, the model must be capable of quickly learning sufficiently general discriminative information from a limited amount of labeled data. 
On the other hand, once it has acquired this discriminative information, the model should effectively leverage the unlabeled data for further optimization.
Current SSL methods mainly focus on the latter aspect, devising strategies to more effectively utilize unlabeled data, such as utilizing the predictions as pseudo labels for supervision~\cite{bai2017semi,fan2020inf}, and imposing a consistency regularization on the predictions of different models or branches~\cite{wu2022mutual,yang2023revisiting,yu2019uncertainty}.
However, the first key aspect of rapid learning from limited labeled data is often overlooked.
To overcome this limitation, we draw our attention to the general segmentation foundation model, i.e., the segment anything model (SAM), which is pre-trained on a large-scale natural datasets and has the potential of transferring to a new task by using only limited labeled data with the impressive few-shot learning capabilities demonstrated in prior research~\cite{chen2023ma}. 

% The key to the success of SSL methods is first helping the segmentation model accumulate general knowledge for segmentation efficiently using the limited labels, such as the discrimination of edges, based on which we can obtain reasonable predictions for unlabeled data, and use them to further optimize the model.
% Current methods tend to focus on the second part and propose various ways to utilize the predictions of unlabeled data, such as utilizing the predictions as pseudo labels for supervision~\cite{bai2017semi,fan2020inf}, and imposing a consistency regularization on the predictions of different models or branches~\cite{wu2022mutual,yang2023revisiting,yu2019uncertainty}.
% Nonetheless, the efficient segmentation knowledge accumulation is often overlooked.

% Aiming to improve the SSL result from this aspect, we notice the segment anything model (SAM)~\cite{kirillov2023segment}.
% Since it is trained on 11M images with over 1 billion masks of different objects, it gains enough general segmentation knowledge and has the potential of transferring to a new task with only limited data.
% With better segmentation knowledge of SAM, we can generate more accurate and reliable predictions for unlabeled data whose importance of SSL has been proved by~\cite{miao2023sc,yao2022enhancing}.
Current methods adapting SAM to medical image segmentation tend to train SAM in a fully supervised way with plenty of labeled data~\cite{wu2023medical,zhang2023customized}.
Very recently, only a limited number of works attempt to adapt SAM in the SSL setting.
% , by using SAM as an auxiliary module to generate more reliable predictions for unlabeled data
For example, Samdsk~\cite{zhang2023samdsk} leverages SAM to produce pseudo labels and select reliable ones into the labeled set to train a traditional segmentation network, i.e., a convolutional neural network (CNN).
% The selection is guided by both the output of the CNN and the image-level constraint like the RoI number.
Li et al.~\cite{li2023segment} generate prompts from the prediction of a CNN and then choose outputs with a high consistency between the CNN and SAM as pseudo labels.
SemiSAM~\cite{zhang2023semisam} produces prompts in a similar way by using the CNN trained in a Mean Teacher framework and uses SAM's output as an additional supervision signal.
% Moreover, a correction model is developed by ~\cite{chen2023aslseg,huang2023push} to further refine SAM's predictions.
In these methods, SAM is simply leveraged as a static and standalone component to generate pseudo labels on medical images, which may not yield desired performance due to the significant domain gap between natural and medical images~\cite{deng2023segment,he2023accuracy}. 
% Despite the promising results, SAM is usually frozen in these methods, which might bottleneck their performance.
% Due to the domain gap between natural images and medical images, directly applying SAM without fine-tuning to specific medical images cannot obtain satisfactory results in some cases even using ground truth labels for prompt generation~\cite{deng2023segment,he2023accuracy}, such as the irregular ring structure of the myocardium segmentation~\cite{miao2023sc}.
Chen et al.~\cite{chen2023aslseg} include the fine-tuning of SAM into the loop of SSL and thus enhance the adaptation ability of SAM to medical images.
However, this work only fine-tunes SAM with a small number of labeled data whereas information contained in the large number of unlabeled data is not fully explored.

In this paper, we aim to leverage the few-shot learning capabilities of the SAM model to bolster our SSL framework for rapid learning from a limited amount of labeled data. Building on this foundation, we then leverage the unique advantage of SAM's prompting mechanism~\cite{kirillov2023segment}, to develop effective strategies for learning from unlabeled data in SSL.
% One of the core designs of SAM is the promptable segmentation~\cite{kirillov2023segment}.
% By providing prompts to specify where or what to segment and combining it with the general knowledge for segmentation, a reasonable prediction can be obtained for a new image.
We propose a semi-supervised medical image segmentation framework which is driven by \textbf{c}ross \textbf{p}rompting \textbf{c}onsistency with \textbf{s}egment \textbf{a}nything \textbf{m}odel (CPC-SAM).
Our method innovates a SAM enabled cross-prompting strategy within a dual-branch framework, which uses the unprompted output from one branch to generate prompts for the other branch. Then the prompted output from the second branch is employed to guide the training of the first branch. Such a cross prompting and supervision strategy enhances the learning process, effectively leveraging the unlabeled data.
{Nonetheless, w}ithout ground truth for the unlabeled inputs, the prompts generated from unprompted outputs can be inherently unreliable and noisy. The prompted output is thus probably to be unreliable as well due to SAM's high sensitivity to prompt positions. To address this issue, we design a novel prompt consistency regularization strategy aimed at improving the consistency of outputs across varying prompts. This strategy reduces SAM's sensitivity to different prompts and enhances the invariance of the output, ensuring more reliable and stable results even when derived from less dependable prompts.
Our method has been extensively evaluated on two public datasets for breast cancer segmentation and cardiac structure segmentation, showing superiority over existing methods, especially when the labeled data are extremely limited.
Specifically, using only 10 labeled ultrasound images, our method obtains an improvement of over 9\% Dice than various strong baselines on the breast cancer segmentation task.

\section{Method}
\begin{figure}[!t]
 \centering
 \includegraphics[width=0.95\textwidth]{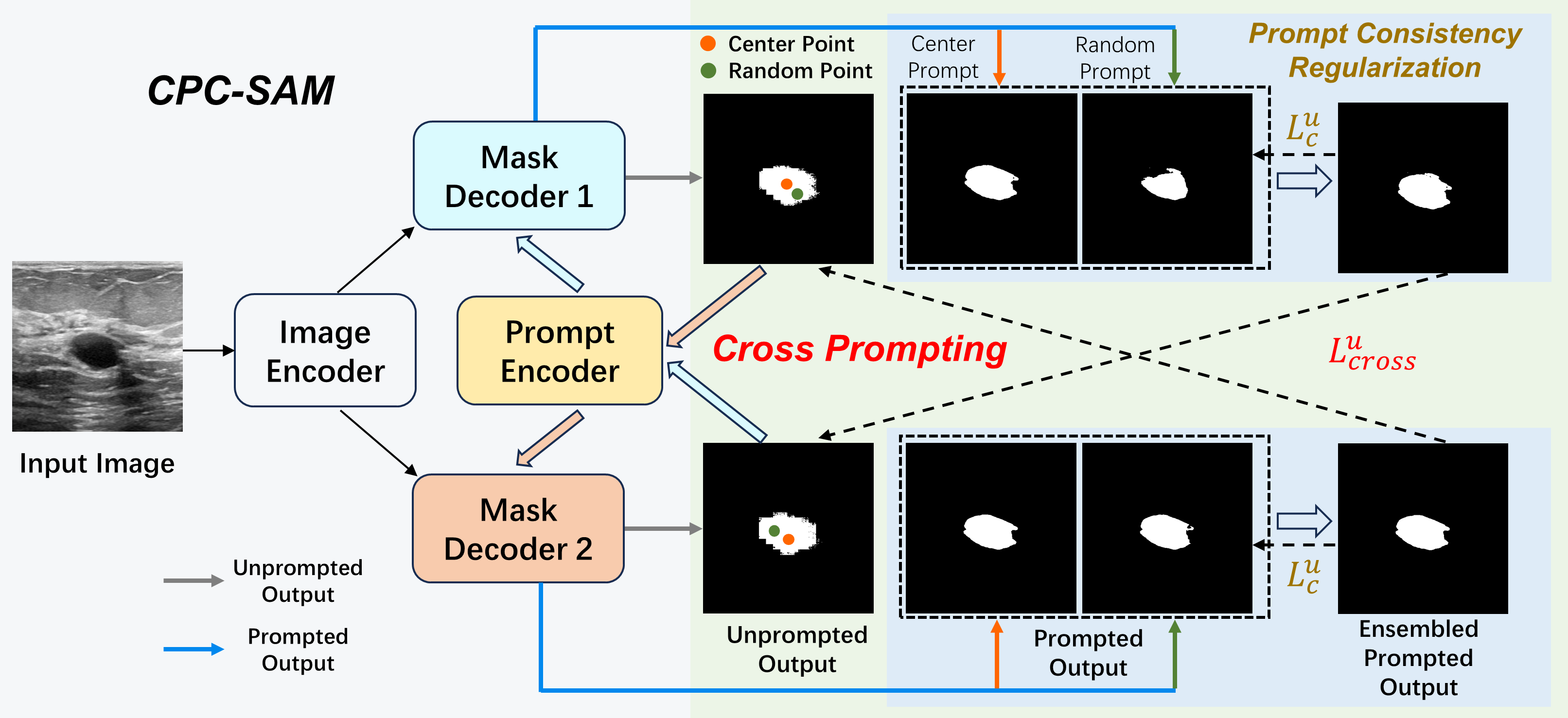}
 \caption{The overview of our proposed method.
 The adapted dual-branch SAM is fine-tuned by the cross prompting loss $L_{cross}^u$ with a prompt consistency regularization $L_c^u$ on the unlabeled data in addition to the supervised loss.} \label{method}
\end{figure}

Fig.~\ref{method} gives an overview of our cross prompting consistency framework with SAM for SSL medical image segmentation, called CPC-SAM.
% In order to reduce the domain gap and improve the prediction quality, we directly fine-tune SAM on target datasets.
Considering the few-shot learning capabilities of the SAM model, we directly fine-tune SAM in the SSL pipeline to achieve the rapid learning from a limited amount of labeled data.
Building on this foundation, a cross prompting dual-branch framework is developed based on the promptable property of SAM to make full use of the large scale of unlabeled data.
Moreover, considering the potential harmfulness of SAM's sensitivity to prompts' positions for SSL, we further propose the prompt consistency regularization to enhance output invariance under various prompts.

\subsection{Problem Formulation and Architecture}
SSL segmentation aims to obtain a satisfactory performance using a small number of labeled data $\mathcal{L} = \left\{ {\left( {{{\boldsymbol{x}}_i},{{\boldsymbol{y}}_i}} \right)} \right\}_{i = 1}^N$ and a large scale of unlabeled data $\mathcal{U} = \left\{ {\left( {{{\boldsymbol{x}}_i}} \right)} \right\}_{i = N + 1}^{N + M}$, where ${{\boldsymbol{x}}_i} \in {\mathcal{R}^{H \times W}}$ indicates an $H \times W$ image and ${{\boldsymbol{y}}_i} \in {\left\{ {0,1} \right\}^{H \times W \times C}}$ denotes the corresponding annotation for labeled data with $C$ semantic classes.
To improve the prediction quality of SAM on the target dataset, we directly fine-tune SAM in the SSL setting using all the available data.
The fine-tuned SAM also functions as the final segmentation model.
To enable the better use of the unlabeled data, we propose a cross prompting strategy introduced later and adapt the original architecture of SAM to a dual-branch SAM with one shared image encoder $\mathcal{E}$ and prompt encoder $\mathcal{P}$, on top of which two decoders $\mathcal{D}_1, \mathcal{D}_2$ with the same structure but different weight initializations are used to encourage the output diversity.
The function of each module remains the same as the original SAM~\cite{kirillov2023segment}, with the image encoder to extract feature embeddings from the image, the prompt encoder to output prompt embeddings for given prompts, and the mask decoder to produce segmentation results based on the feature embeddings and prompt embeddings.
As done in~\cite{zhang2023customized}, when no explicit prompts are given, the default dense prompt embedding is used and fine-tuned during training for automatic segmentation, which is also used during inference.

% \begin{figure}[!t]
%  \centering
%  \includegraphics[width=0.95\textwidth]{Figures/method_20240308.png}
%  \caption{The overview of our proposed method.
%  The adapted dual-branch SAM is fine-tuned by the cross prompting loss $L_{cross}^u$ with a prompt consistency regularization $L_c^u$ on the unlabeled data in addition to the supervised loss.} \label{method}
% \end{figure}

% In this way, we can obtain the segmentation result using the input image alone without any explicit prompts during inference.
% Then, we use the LoRA strategy~\cite{hu2021lora} to efficiently fine-tune the the most heavy part, i.e., the image encoder following~\cite{zhang2023customized}.

% \subsection{Cross Prompting for Cross Pseudo Supervision}
\subsection{SAM-enabled Cross Prompting}
Although fine-tuning on the target dataset can effectively integrate domain knowledge of specific medical images to SAM, current methods only use the small labeled set for fine-tuning~\cite{chen2023aslseg}, neglecting the potential of the large number of unlabeled images in the SSL setting and thus limiting the fine-tuning performance on target datasets.
Therefore, we propose a cross prompting scheme based on the promptable property of SAM to make full use of the unlabeled data and integrate more domain knowledge on the target task to SAM.

First, we generate prompts for each other from the unprompted outputs under our dual-branch framework.
% Intuitively, the output of SAM given an appropriate prompt $\hat p^u$ should be more accurate and reliable compared to the output without any explicit prompts $p^u$, since the prompt offers position information of the target area.
% Based on this intuition, our cross prompting scheme generates prompts for each other and use the prompted output to guide the optimization of unprompted predictions for unlabeled data under our dual-branch framework.
Here, point prompt is used following~\cite{chen2023aslseg,huang2023push,li2023segment,zhang2023semisam} for its simplicity and flexibility.
Take the first branch $\mathcal{D}_1$ for prompt generation as an example.
Given an unlabeled image $\boldsymbol{x} \in \mathcal{U}$, we first obtain the output $p_1^u$ from $\mathcal{D}_1$ using the feature embedding and the default prompt embedding since no explicit prompt is provided: $p_1^u = \mathcal{D}_1(\mathcal{E}(x), \mathcal{P}(None))$.
Then, we generate the point prompt $Pt_2$ from $p_1^u$ by selecting the center or a random point of the largest connected component of the object of interest.
After that, we can obtain the prompted output of $\mathcal{D}_2$: $\hat p_{2}^u = \mathcal{D}_2(\mathcal{E}(x), \mathcal{P}(Pt_2))$.
Similarly, we can obtain the unprompted output $p_2^u$ when using $\mathcal{D}_2$ to produce prompts and the prompted prediction $\hat p_{1}^u$ output by $\mathcal{D}_1$.
Second, we use the prompted outputs to guide the optimization of unprompted predictions based on the intuition that the output of SAM given an appropriate prompt $\hat p^u$ should be more accurate and reliable compared to the output without any explicit prompts $p^u$, since the prompt offers position information of the target area.
To alleviate the confirmation bias of using the same branch to supervise itself, the prompted output $\hat p_{1}^u$ of $\mathcal{D}_1$ is used to supervise the unprompted prediction $p_{2}^u$ of $\mathcal{D}_2$, and vice versa.
Therefore, our cross prompting loss is a symmetrical constraint with a combination of the Dice loss and the cross-entropy loss:
\begin{equation}
\label{L_{cross}}
{L_{cross}^u} = \frac{1}{2}\left[ {{L_{dice}}(p_{1}^u, \hat p_{2}^u) + {L_{ce}}(p_{1}^u, \hat p_{2}^u)} \right] + \frac{1}{2}\left[ {{L_{dice}}(p_{2}^u, \hat p_{1}^u) + {L_{ce}}(p_{2}^u, \hat p_{1}^u)} \right]
\end{equation}
% The effectiveness of such an intuition has been verified by EviPrompt~\cite{xu2023eviprompt}, which uses the prompts generated from the coarse mask as inputs for a second forward process and obtain a refined mask.
% Howevere, the SAM in~\cite{xu2023eviprompt} is frozen, potentially limiting the performance on challenging tasks due to the domain gap.
% In this work, we use such a strategy to include unlabeled data for fine-tuning by generating prompts for each other and use the prompted output to guide the optimization of unprompted predictions under our dual-branch framework, forming a self-boosting loop for performance improvement.
Compared with the vanilla cross pseudo supervision method that directly uses the output to supervise each other~\cite{chen2021semi}, our cross prompting scheme makes full use of the promptable property of SAM as a refinement step to obtain a better pseudo label as a better guidance than the unprompted output.
The final SSL performance is thus improved as shown in the ablation studies.

\subsection{Prompt Consistency Regularization}
The cross prompting scheme solves how to utilize unlabeled images to improve the fine-tuning performance.
The key to its success is generating reliable predictions in the dilemma where SAM's output is sensitive to the prompt locations whereas the point prompts generated from the noisy coarse mask in SSL tend to have a high variance and a low accuracy in terms of positions.
The core of alleviating this dilemma is to enhance the output invariance under various prompts.
% In this way, we can loose the requirement of prompt position and are more likely to obtain reliable outputs in the SSL setting since more positions on the target area can be selected as a prompt.
% Empirically, the center point tend to provide a reasonable result but a point near the edge has a high chance to mis-classify the background as the foreground.
In the ideal case, predictions under two different prompts on the target area should be the same and approach the ground truth as close as possible in the meanwhile.
Based on this motivation, we propose a novel prompt consistency regularization (PCR) loss to enhance the output invariance of SAM under various prompts.
Take the prompted outputs of $\mathcal{D}_1$ as an example, where $\mathcal{D}_2$ is used to generate prompts, we first find the largest connected component of each semantic class of the unprompted prediction $p_{2}^u$ as a post-processing step for the noisy output to increase the chance of selecting a point on the hidden ground truth area.
After that, a center point and a random point {are} selected from the the largest connected component, based on which two seperate predictions are obtained, denoted as $\hat p_{1,c}^u,\hat p_{1,r}^u$, respectively.
The center point is selected since it can provide a stable prediction empirically, while the random point prompt is chosen to simulate the potential prompt variance in SSL segmentation.
Then, $\hat p_{1,r}^u$ should be similar to $\hat p_{1,c}^u$.
Since the ensemble of multiple predictions can usually obtain a more robust result, we use the ensemble of both predictions $\hat p_{1}^u=1/2(\hat p_{1,r}^u + \hat p_{1,c}^u)$ as a more reliable guidance for the randomly prompted prediction $\hat p_{1,r}^u$.
Symmetrically, we can obtain the prompted outputs $\hat p_{2,c}^u,\hat p_{2,r}^u$ and the ensemble result $\hat p_{2}^u$ of $\mathcal{D}_2$. 
The prompt consistency regularization is applied to both decoders:
\begin{equation}
\label{L_{c}}
L_{c}^u = \frac{1}{2}\left[ {{L_{dice}}(\hat p_{1,r}^u, \hat p_{1}^u) + {L_{ce}}(\hat p_{1,r}^u, \hat p_{1}^u)} \right] + \frac{1}{2}\left[ {{L_{dice}}(\hat p_{2,r}^u, \hat p_{2}^u) + {L_{ce}}(\hat p_{2,r}^u, \hat p_{2}^u)} \right]
\end{equation}
% where $\hat p_{1,r}^u,\hat p_{2,r}^u$ are the predictions of the two mask decoders using the random prompts and $\hat p_{i}^u=1/2(\hat p_{i,r}^u + \hat p_{i,c}^u), i=1,2$ represent the ensemble results of the random prompt and center prompt.
The ensemble results $\hat p_{1}^u,\hat p_{2}^u$ are also used to be the supervision signals in Equ.~\ref{L_{cross}} as a by-product of PCR.
Also, since the output using the center prompt tends to be more stable, the PCR is only applied to $p_{1,r}^u,p_{2,r}^u$.
The efficacy of such a design will be proved in ablation studies.

Meanwhile, to ensure the prompted outputs can approach the ground truth well, we supervise the prompted outputs $p_{1,c}^l,p_{2,c}^l,p_{1,r}^l,p_{2,r}^l$ with annotations on the labeled set besides the supervised loss on unprompted outputs $p_{1}^l,p_{2}^l$ in SSL:
\begin{equation}
\label{L_s}
{L_s} = L_s^l(p_1^l, \boldsymbol{y}) + L_s^l(p_2^l, \boldsymbol{y}) + L_{s,p}^l(p_{1,c}^l, \boldsymbol{y}) + L_{s,p}^l(p_{2,c}^l, \boldsymbol{y}) + L_{s,p}^l(p_{1,r}^l, \boldsymbol{y}) + L_{s,p}^l(p_{2,r}^l, \boldsymbol{y})
\end{equation}
where $L_s^l = 0.8L_{dice}+0.2L_{ce}$ following SAMed~\cite{zhang2023customized} and $L_{s,p}^l = 0.5L_{dice}+0.5L_{ce}$.
Finally, the total loss for training is the combination of the supervised loss ${L_s}$ on labeled data, the cross prompting loss ${L_{cross}^u}$ and the prompt consistency regularization loss $L_{c}^u$ on unlabeled data: ${L_{total}} = L_s + \lambda_{1} {L_{cross}^u} + \lambda_{2} L_{c}^u$.

\section{Experimental Results}
\textbf{Datasets.} We evaluate our proposed method on two publicly available datasets: the BUSI dataset~\cite{al2020dataset} and the ACDC dataset~\cite{bernard2018deep}.
The BUSI dataset~\cite{al2020dataset} consists of 647 ultrasound images for breast cancer segmentation, with 437 benign cases and 210 malignant ones.
We randomly split the data on each category and finally obtain 431, 86, and 130 images for training, validation, and testing, respectively.
The ACDC dataset~\cite{bernard2018deep} contains 200 cine MRI scans from 100 patients with three regions of interest, i.e., the right ventricle cavity, the myocardium, and the left ventricle cavity.
Following~\cite{bai2023bidirectional,chen2023decoupled}, the dataset is randomly split on the patient level, with 70 patients for training, 10 for validation, and 20 for testing.

\noindent \textbf{Implementation Details and Evaluation Metrics.} Our method is implemented by Pytorch and trained on an NVIDIA A40 GPU.
Most training settings are the same on both datasets.
Specifically, the ViT\_B version of SAM is employed.
Following~\cite{zhang2023customized}, we apply LoRA~\cite{hu2021lora} to the query and value heads in each transformer block of $\mathcal{E}$ with $r=4$ and optimize all the parameters in $\mathcal{P}$ and $\mathcal{D}_1,\mathcal{D}_2$ through the normal back propagation.
We load the pre-trained weights for the image encoder and prompt encoder, whereas two decoders are initialized randomly.
We expand the number of point prompt embeddings in the prompt encoder to the number of semantic classes for multi-class segmentation.
Also, we increase the output resolution of mask decoders by the progressive upsampling strategy in~\cite{chen2023ma}.
The input images are resized to 512$\times$512 and normalized to [0,1].
Data augmentation used in training include random rotation between [-20$^{\circ}$, 20$^{\circ}$] and random flips.
Our adapted SAM is optimized by an AdamW optimizer for 10000 epochs.
The same warmup and exponential learning rate decay strategy as~\cite{zhang2023customized} are adopted, setting the maximum learning rate as 0.001 and the warmup period as 5000 iterations.
$\lambda_{1}$ and $\lambda_{2}$ are empirically set as 0.4 and 0.05.
The batch size is 6 and 12 for the BUSI and ACDC dataset, respectively, each containing half labeled data.
As done in~\cite{bai2023bidirectional,chen2023decoupled}, four evaluation metrics are taken, including the Dice similarity coefficient (DSC), Jaccard (JC), 95\% Hausdorff Distance (95HD), and the average surface distance (ASD).
The unit of DSC and JC is \%. The unit of HD95 and ASD is pixel and mm on the BUSI and ACDC dataset, respectively, since the resolution is not provided on the BUSI dataset.

\begin{table}[!t]
\caption{{Comparisons with SOTA methods on the BUSI and ACDC dataset. Column "\#Lab" denotes the number of labeled data and the number of all training data, respectively.}}
\label{compare}
\centering
% \tiny
\scriptsize
\resizebox{1.0\linewidth}{!}{
\begin{tabular}{c|c|c|c|c|c||c|c|c|c|c}
\toprule[1.0pt]
\multirow{2}{*}{Method} & \multicolumn{5}{c||}{\textbf{BUSI}}                                                                                                             & \multicolumn{5}{c}{\textbf{ACDC}}                                                                                                       \\ \cline{2-11} 
                        & {\#Lab}        & {DSC↑} & {JC↑}   & {HD95↓}  & ASD↓  & {\#Lab} & {DSC↑} & {JC↑}   & {HD95↓}  & ASD↓  \\ \hline
                        \hline
U-Net~\cite{ronneberger2015u}                    & 431/431 & {77.19} & {68.29} & {75.03}  & 31.21 & 70/70     & {91.53} & {84.77} & {4.23}   & 1.11  \\ \hline
SAM-point(MIA'23)~\cite{mazurowski2023segment}               & {0/431}     & {52.99} & {44.51} & {168.26} & 91.78 & {0/70}      & {62.88} & {49.53} & {20.46}  & 7.07  \\ \hline
U-Net~\cite{ronneberger2015u}                    & {10/431}     & {31.63} & {24.52} & {159.49} & 63.43 & {1/70}      & {29.37} & {20.53} & {107.51} & 52.84 \\
SAMed~\cite{zhang2023customized}                   & {10/431}     & {65.09} & {54.78} & {119.75} & 47.84 & {1/70}      & {75.01} & {61.53} & {28.99}  & 9.13  \\
UAMT(MICCAI’19)~\cite{yu2019uncertainty}        & {10/431}     & {40.93} & {30.96} & {175.31} & 76.51 & {1/70}      & {29.14} & {20.14} & {107.69} & 53.58 \\
CPS(CVPR’21)~\cite{chen2021semi}           & {10/431}     & {32.92} & {25.70} & {144.92} & 50.54 & {1/70}      & {30.46} & {21.00} & {95.74}  & 45.48 \\
URPC(MIA’22)~\cite{luo2022semi}           & {10/431}     & {32.16} & {24.75} & {151.59} & 64.97 & {1/70}      & {31.00} & {20.81} & {123.03} & 59.94 \\
MC-Net+(MIA’22)~\cite{wu2022mutual}        & {10/431}     & {36.24} & {27.45} & {167.91} & 71.80 & {1/70}      & {38.84} & {28.58} & {62.21}  & 30.67 \\
DCNet(MICCAI’23)~\cite{chen2023decoupled}       & {10/431}     & {42.14} & {32.11} & {154.39} & 64.21 & {1/70}      & {41.13} & {31.61} & {56.16}  & 24.71 \\
BCP(CVPR’23)~\cite{bai2023bidirectional}           & {10/431}     & {61.81} & {51.12} & {112.91} & {38.15} & {1/70}      & {68.39} & {56.8}  & {50.9}   & 21.99 \\
UniMatch(CVPR’23)~\cite{yang2023revisiting}      & {10/431}     & {60.98} & {49.85} & {109.79} & 47.50 & {1/70}      & {84.47} & {74.25} & {15.36}  & 4.57  \\
SemiSAM~\cite{zhang2023semisam}                 & {10/431}     & {43.43} & {32.48} & {177.30} & 84.46 & {1/70}      & {34.18} & {23.96} & {100.75} & 47.03 \\
CPC-SAM (ours)                    & {10/431}     & {\textbf{71.20}} & {\textbf{61.15}} & {\textbf{100.22}} & \textbf{37.86} & {1/70}      & {\textbf{85.56}} & {\textbf{75.74}} & {\textbf{9.19}}   & \textbf{2.84}  \\ \hline
U-Net~\cite{ronneberger2015u}                    & {20/431}   & {44.22} & {34.73} & {160.04} & 69.52 & {3/70}      & {45.95} & {35.96} & {71.11}  & 32.47 \\
SAMed~\cite{zhang2023customized}                   & {20/431}   & {67.28} & {57.55} & {107.31} & 49.70 & {3/70}      & {83.04} & {71.98} & {14.93}  & 4.05  \\
UAMT(MICCAI’19)~\cite{yu2019uncertainty}        & {20/431}   & {45.83} & {35.84} & {163.53} & 80.92 & {3/70}      & {56.67} & {45.93} & {15.06}  & 45.24 \\
CPS(CVPR’21)~\cite{chen2021semi}           & {20/431}   & {46.74} & {37.61} & {142.73} & 56.70 & {3/70}      & {56.87} & {46.88} & {20.18}  & 2.91  \\
URPC(MIA’22)~\cite{luo2022semi}           & {20/431}   & {45.26} & {35.51} & {173.11} & 73.47 & {3/70}      & {55.98} & {44.75} & {40.47}  & 14.13 \\
MC-Net+(MIA’22)~\cite{wu2022mutual}        & {20/431}   & {47.29} & {33.00} & {183.14} & 84.53 & {3/70}      & {65.37} & {54.18} & {27.64}  & 6.32  \\
DCNet(MICCAI’23)~\cite{chen2023decoupled}       & {20/431}   & {56.87} & {46.60} & {130.31} & 56.14 & {3/70}      & {72.21} & {62.27} & {26.50}  & 10.59 \\
BCP(CVPR’23)~\cite{bai2023bidirectional}           & {20/431}   & {65.54} & {56.05} & \textbf{93.07}  & \textbf{39.09} & {3/70}      & {87.57} & {78.58} & {8.68}   & 2.30  \\
UniMatch(CVPR’23)~\cite{yang2023revisiting}      & {20/431}   & {62.47} & {51.48} & {100.73} & 45.88 & {3/70}      & {87.31} & {78.20} & {8.62}   & 2.74  \\
SemiSAM~\cite{zhang2023semisam}                 & {20/431}   & {50.09} & {38.63} & {170.42} & 77.85 & {3/70}      & {51.01} & {39.45} & {70.13}  & 28.26 \\
CPC-SAM (ours)                    & {20/431}   & \textbf{72.41} & \textbf{62.72} & {96.26} & 40.93 & {3/70}      & \textbf{87.95} & \textbf{79.01} & \textbf{5.80}    & \textbf{1.54}  \\ \toprule[1.0pt]
\end{tabular}
}
\end{table}

\begin{figure}[!t]
 \centering
 \includegraphics[width=0.9\textwidth]{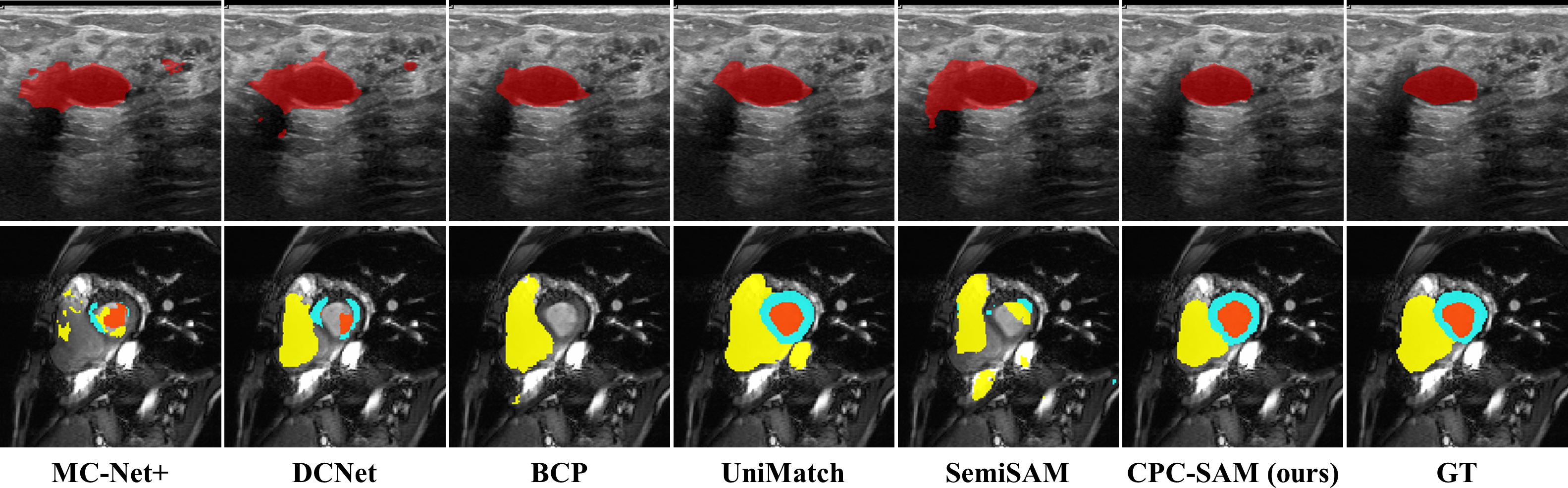}
 \caption{Visualizations of different methods on the BUSI dataset with 10 labeled images (top) and on the ACDC dataset with 1 labeled patient (bottom).} \label{visual}
\end{figure}

% \subsection{Comparisons with the State-of-the-arts}

\noindent \textbf{Comparisons with the State-of-the-arts.} We compare our method with the state-of-the-arts (SOTA) SSL methods, including UAMT~\cite{yu2019uncertainty}, CPS~\cite{chen2021semi}, URPC~\cite{luo2022semi}, MC-Net+~\cite{wu2022mutual}, DCNet~\cite{chen2023decoupled}, BCP~\cite{bai2023bidirectional}, and UniMatch~\cite{yang2023revisiting}, and a representative SAM-based SSL method SemiSAM~\cite{zhang2023semisam}.
We also compare with the supervised counterparts trained with labeled data alone, i.e., U-Net~\cite{ronneberger2015u} and SAMed~\cite{zhang2023customized}.
% The U-Net counterpart using all the annotations is considered as the upper bound.
The zero-shot performance of SAM is also included using the center point prompt generated from the ground truth labels following~\cite{mazurowski2023segment}, denoted as "SAM-point".
As shown in Table~\ref{compare}, our method significantly outperforms other SSL methods on the BUSI dataset by a margin of over 9.3\% and 6.8\% DSC when different numbers of labeled data are used.
On the ACDC dataset, our method obtains the best results on all the metrics across the two labeled-data ratios.
% Our result is superior to other SSL methods by about 1.1\% and 0.4\% DSC when the labels of 1 and 3 patients are used, respectively.
It is worth noting that although SAMed only using the labeled data can obtain comparable and even better results to some SSL methods, our proposed method obtains significantly better results on both datasets thanks to the effective use of unlabeled data.
Moreover, by using SAM to generate predictions for unlabeled data, SemiSAM obtains a superior result to most SSL methods on the BUSI dataset.
However, its performance on the ACDC dataset degrades steeply, probably due to the difficulty of segmenting the ring structure of the myocardium without fine-tuning as shown in~\cite{miao2023sc}.
The robustness and superior results of our method validate the necessity of fine-tuning on the target dataset and effectiveness of our proposed method.
The ensemble results of two branches is reported in this work, but the difference between each branch and the ensembled result is marginal, e.g., 85.59\%, 85.47\% and 85.56\% DSC for $\mathcal{D}_1, \mathcal{D}_2$, and ensemble, respectively on the ACDC dataset with 1 labeled patient.
The visualization comparisons in Fig.~\ref{visual} further validate the superiority of our method.

\begin{figure}[!tbp]
\begin{minipage}[b]{.65\linewidth}
\centering
% \small
\scriptsize
% \label{ablation}
\begin{tabular}{c|c|c|c|c|c|c}
\hline
SSL & Cross Prompting & PCR & DSC↑ & JC↑ & HD95↓ & ASD↓ \\ \hline
    &       &                      & 77.26            & 64.88         & 15.72           & 5.22          \\
 \checkmark  &              &                      & 80.20       & 68.44    & 19.48      & 6.52      \\
\checkmark   & \checkmark            &                      & 84.75           &  74.39        &  13.41          &  3.84         \\
 \checkmark   & \checkmark            & \checkmark                    & \textbf{85.56}      & \textbf{75.74}    & \textbf{9.19}       & \textbf{2.84}      \\
\hline
\end{tabular}
\captionof{table}{Ablation studies of different components of our method on the ACDC dataset.}
\label{tab:ablation}
\end{minipage}
\begin{minipage}[b]{.3\linewidth}
\centering
\includegraphics[width=0.8\textwidth, height=0.4\textwidth]{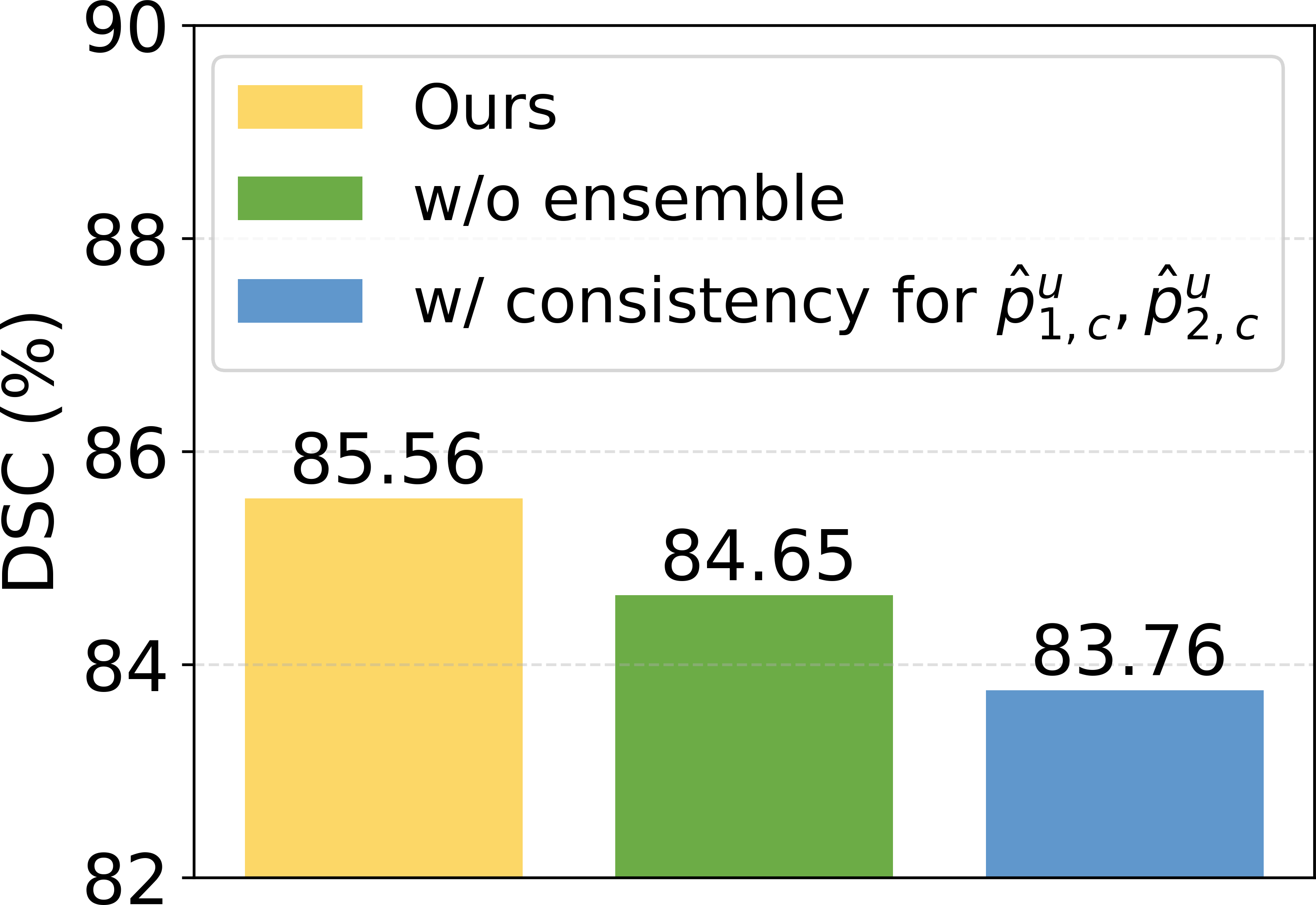}
\caption{Effects of various consistency constraints.}
\label{fig:consistency}
\end{minipage}
\end{figure}

\begin{figure}[!tbp]
\begin{minipage}[b]{.65\linewidth}
\centering
% \small
\scriptsize
% \label{ablation}
\begin{tabular}{c|c|c|c|c|c}
\hline
\# Center & \# Random & DSC↑ & JC↑ & HD95↓ & ASD↓ \\ \hline
    0  & 2                     & 84.95            & 74.66         & 12.53           & 3.64          \\
      1        &      1                & \textbf{85.56}      & \textbf{75.74}    & \textbf{9.19}       & \textbf{2.84}      \\
 1           &       5               & 84.46           &  74.10        &  13.19          &  3.63         \\
 1            & 10                    & 84.23      & 73.61    & 12.75       & 3.95      \\
\hline
\end{tabular}
\captionof{table}{Effects of different numbers of center and random points in the PCR on the ACDC dataset.}
\label{tab:number}
\end{minipage}
\begin{minipage}[b]{.3\linewidth}
\centering
\includegraphics[width=0.8\textwidth, height=0.4\textwidth]{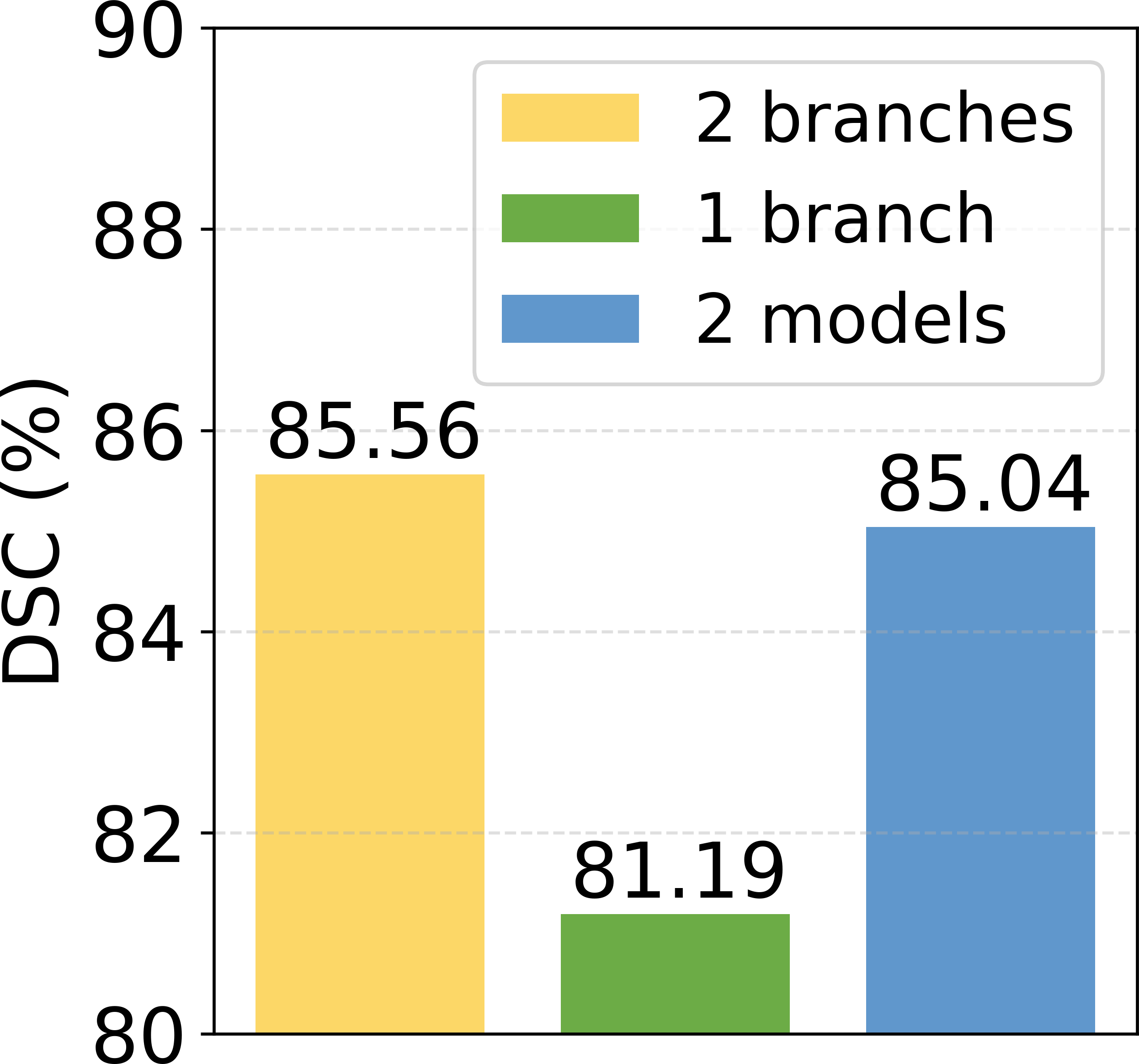}
\caption{Effects of different architectures.}
\label{fig:architect}
\end{minipage}
\end{figure}

% \subsection{Ablation Studies}

\noindent \textbf{Ablation Studies.}
Table~\ref{tab:ablation} shows ablation studies on the key components of our method with the ACDC dataset.
Obviously, introducing unlabeled data into fine-tuning (Row 2-4) significantly outperforms training on the labeled data alone (Row 1) with a margin of over 2.9\% DSC.
Also, using only the center point prompt (Row 3) for cross prompting, we can obtain superior results over the method using the unprompted outputs to supervise each other (Row 2).
With the help of PCR, the result is further improved by 0.81\% DSC.
Also, we validate the necessity of using ensemble results and dropping the regularization on the center prompted outputs in Fig.~\ref{fig:consistency}.
We further validate the efficacy of our center-random prompt selection strategy {(Row 2)} in Table~\ref{tab:number}.
{I}ntroducing more random points leads to a lower performance, possibly because more random point prompts can include some points outside the hidden ground truth for unlabeled data.
Moreover, we apply our method to other architectures, such as the original SAM with only one mask decoder and 2 models with two SAMs (See Fig.~\ref{fig:architect}).
The inferior result using 1 branch might be caused by the more serious confirmation bias problem in such a self-training framework~\cite{chen2021semi,ke2019dual}.
% We also explore the potential of using more random point prompts in our methods, and obtain a DSC 84.46\% and 84.23\%, which are lower than using only one random point prompt (85.56\%), possibly because more random point prompts can include some points outside the hidden ground truth considering the noisy predictions in the SSL setting.
% The ensemble results of two branches is reported in this work, but the difference between each branch and the ensembled result is marginal, e.g., 85.59\%, 85.47\% and 85.56\% DSC for $\mathcal{D}_1, \mathcal{D}_2$, and ensemble, respectively on the ACDC dataset with 1 labeled patient.

\section{Conclusion}
This paper proposes a cross prompting framework to adapt SAM for SSL medical image segmentation.
A novel cross prompting scheme with a prompt consistency regularization is developed to make full use of the unlabeled data for performance improvement under the challenges of domain gap and prompt sensitivity.
Comparisons on two datasets demonstrate the efficacy of our method, especially when labeled data are extremely limited.
In the future, we'll explore more strategies to select appropriate prompts for reliable outputs and explore the potential of using other kinds or modalities of prompts.

\begin{credits}
\subsubsection{\ackname} The work described in this paper was supported in part by the Research Grants Council of the Hong Kong Special Administrative Region, China, under Project T45-401/22-N; and by the Hong Kong Innovation and Technology Fund (Project No. GHP/080/20SZ).

% \subsubsection{\discintname}
% The authors have no competing interests to declare.
\end{credits}
%
% ---- Bibliography ----
%
% BibTeX users should specify bibliography style 'splncs04'.
% References will then be sorted and formatted in the correct style.
%
% \bibliographystyle{splncs04}
% \bibliography{mybibliography}
%
\bibliographystyle{splncs04}
\bibliography{Paper-0321}
\end{document}